\documentclass[10pt,twocolumn,letterpaper]{article}

\usepackage{cvpr}
\usepackage{times}
\usepackage{epsfig}
\usepackage{graphicx}
\usepackage{amsmath}
\usepackage{amssymb}

\usepackage{mathtools}

\newcommand{\norm}[1]{\left\lVert#1\right\rVert}
\usepackage[inline]{enumitem}
\usepackage{booktabs}

\usepackage{multirow}
\usepackage{verbatim}

\usepackage[breaklinks=true,bookmarks=false]{hyperref}

\cvprfinalcopy % *** Uncomment this line for the final submission

\def\cvprPaperID{4060} % *** Enter the CVPR Paper ID here

\ifcvprfinal\fi
\begin{document}

%%%%%%%%% TITLE
\title{IPOD: Intensive Point-based Object Detector for Point Cloud}

\author{Zetong Yang$^{\dag}$\\
\and
Yanan Sun$^{\dag}$
\and
Shu Liu$^{\dag}$
\and 
Xiaoyong Shen$^{\dag}$
\and 
Jiaya Jia$^{\dag, \ddagger}$
\\
\and
$^{\dag}$Youtu Lab, Tencent~~~~~~$^{\ddagger}$The Chinese University of Hong Kong\\
\vspace{-2mm}
{\tt\small \{tomztyang, now.syn, liushuhust, Goodshenxy\}@gmail.com~~leojia@cse.cuhk.edu.hk} 
}
\maketitle
%\thispagestyle{empty}

%%%%%%%%% ABSTRACT
\begin{abstract}
   We present a novel 3D object detection framework, named IPOD, based on raw point cloud. It seeds object proposal for each point, which is the basic element. This paradigm provides us with high recall and high fidelity of information, leading to a suitable way to process point cloud data. We design an end-to-end trainable architecture, where features of all points within a proposal are extracted from the backbone network and achieve a proposal feature for final bounding inference. These features with both context information and precise point cloud coordinates yield improved performance. We conduct experiments on KITTI dataset, evaluating our performance in terms of 3D object detection, Bird's Eye View (BEV) detection and 2D object detection. Our method accomplishes new state-of-the-art , showing great advantage on the hard set.
\end{abstract}

\begin{figure*}[bpt]
\centering
\includegraphics[width=1.0\linewidth]{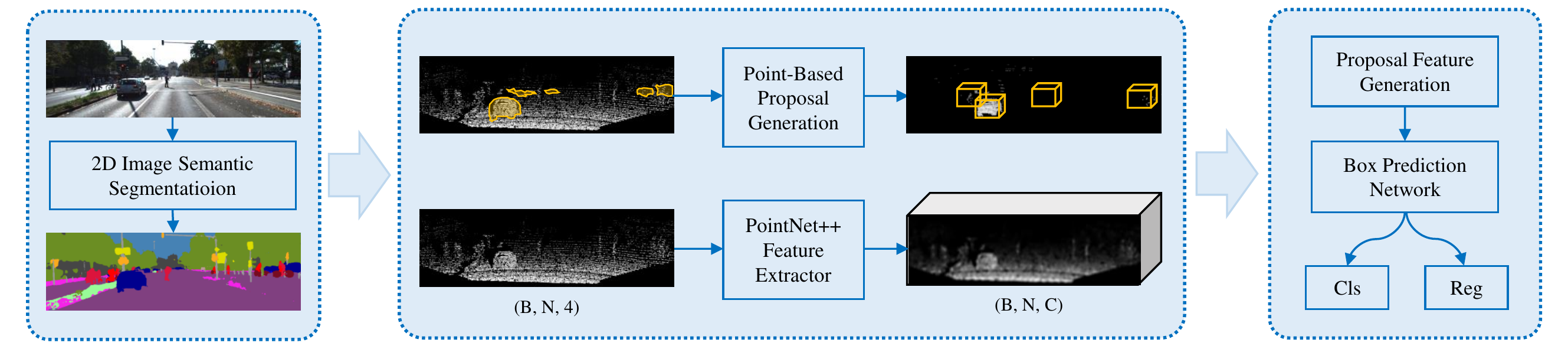}\\
\caption{Illustration of our framework. It consists of three different parts. The first is a subsampling network to filter out most background points. The second part is for point-based proposal generation. The third component is the network architecture, which is composed of backbone network, proposal feature generation module and a box prediction network. It classifies and regresses generated proposals.}
\label{fig:framework}
\end{figure*}

%%%%%%%%% BODY TEXT
\section{Introduction}

Great breakthrough has been made in 2D image recognition tasks \cite{RCNN,SSD,FocalLoss,DEFORMABLECONV} with the development of Convolutional Neural Networks (CNNs). Meanwhile, 3D scene understanding with point cloud also becomes an important topic, since it can benefit many applications, such as autonomous driving \cite{KITTIDATASET2} and augmented reality \cite{Multiple3Dtracking}. In this work, we focus on one of the most important 3D scene recognition tasks -- that is, 3D object detection based on point cloud, which predicts the 3D bounding box and class label for each object in the scene.

\paragraph{Challenges} Different from RGB images, LiDAR point cloud is with its unique properties. On the one hand, they provide more spatial and structural information including precise depth and relative location. On the other hand, they are sparse, unordered, and even not uniformly distributed, bringing huge challenge to 3D recognition tasks. 

To deploy CNNs, most existing methods convert 3D point clouds to images by projection \cite{MV3D,AVOD,MultiViewRandomForest,PedestrianDetectionCombine,Vote3Deep} or voxelize cloud with a fixed grid \cite{VoxNet,VotetoVote,VOXELNET}. 
With the compact representation, CNN is applied. Nevertheless, these hand-crafted representations may not be always optimal regarding the detection performance. Along another line, F-PointNet \cite{FPOINTNET} crops point cloud in a frustum determined by 2D object detection results. Then a PointNet \cite{POINTNET} is applied on each frustum to produce 3D results. The performance of this pipeline heavily relies on the image detection results. Moreover, it is easily influenced by large occlusion and clutter objects, which is the general weakness of 2D object detectors. 

\paragraph{Our Contribution} To address aforementioned drawbacks, we propose a new paradigm based on raw point cloud for 3D object detection. We take each point in the cloud as the element and seed them with object proposals. The raw point cloud, without any approximation, is taken as input to keep sufficient information. This design is general and fundamental for point cloud data, and is able to handle occlusion and clutter scenes. 

We note it is nontrivial to come up with such a solution due to the well-known challenges of heavily redundant proposals and ambiguity on assigning corresponding ground-truth labels. Our novelty is on a proposal generation module to output proposals based on each point and effective selection of representative object proposals with corresponding ground-truth labels to ease network training. 
Accordingly, the new structure extracts both context and local information for each proposal, which is then fed to a tiny PointNet to infer final results. 

We evaluate our model on 2D detection, Bird's Eye View (BEV) detection, and 3D detection tasks on KITTI benchmark \cite{KITTI3DBENCHMARK}. Experiments show that our model outperforms state-of-the-art LIDAR based 3D object detection frameworks especially for difficult examples. Our experiments also surprisingly achieve extremely high recall without the common projection operations. Our primary contribution is manifold.
\begin{itemize}

\item We propose a new proposal generation paradigm for point cloud based object detector. It is a natural and general design, which does not need image detection while yielding much higher recall compared with widely used voxel and projection-based methods.

\item A network structure with input of raw point cloud is proposed to produce features with both context and local information.

\item Experiments on KITTI datasets show that our framework better handles many hard cases with highly occluded and crowded objects, and achieves new state-of-the-art performance.
\end{itemize}

% Related work
\section{Related Work}
\paragraph{3D Semantic Segmentation} There have been several approaches to tackle semantic segmentation on point cloud. In \cite{SQUEEZESEG}, a projection function converts LIDAR points to a UV map, which is then classified by 2D semantic segmentation \cite{SQUEEZESEG,PSPNET,DEEPLAB} in pixel level. In \cite{3DMV,SCANNET}, a multi-view based function produces the segmentation mask. The method fuses  information from different views. Other solutions, such as \cite{POINTNET2,POINTNET,POINTCNN,POINTSIFT,SONET}, segment the point cloud from raw LIDAR data. They directly generate features on each point while keeping original structural information. Specifically, a max-pooling method gathers the global feature; it is then concatenated with local feature for processing.

\paragraph{3D Object Detection} There are roughly three different lines for 3D object detection. They are voxel-grid based, multi-view based and PointNet based methods.

$Voxel\textrm{-}grid$ Method: There are several LIDAR-data based 3D object detection frameworks using voxel-grid representation. In \cite{VotetoVote}, each non-empty voxel is encoded with 6 statistical quantities by the points within this voxel. A binary encoding is used in \cite{FullyConvolutionNetworkForVehicle} for each voxel grid. They utilized hand-crafted representation. VoxelNet \cite{VOXELNET} instead stacks many VFE layers to generate machine-learned representation for each voxel. 

$Multi\textrm{-}view$ Method: MV3D \cite{MV3D} projected LIDAR point cloud to BEV and trained a Region Proposal Network (RPN) to generate positive proposals. Afterwards, it merged features from BEV, image view and front view in order to generate refined 3D bounding boxes. AVOD \cite{AVOD} improved MV3D by fusing image and BEV features like \cite{FPN}. Unlike MV3D, which only merges features in the refinement phase, it also merges features from multiple views in the RPN phase to generate more accurate positive proposals. However, these methods still have the limitation when detecting small objects such as pedestrians and cyclists. They do not handle several cases that have multiple objects in depth direction.

$PointNet$ Method: F-PointNet \cite{FPOINTNET} is the first method of utilizing raw point cloud to predict 3D objects. Initially, a 2D object detection module \cite{DSSD} is applied to generate frustum proposals. Then it crops points and passes them into an instance segmentation module. At last, it regresses 3D bounding boxes by the positive points output from the segmentation module. Final performance heavily relies on the detection results from the 2D object detector. In contrast, our design is general and effective to utilize the strong representation power of point cloud.
 
\section{Our Framework}
Our method aims to regress the 3D object bounding box from the easy-to-get point-based object proposals, which is a natural design for point cloud based object detection. To make it feasible, we design new strategies to reduce redundancy and ambiguity existing introduced by trivially seeding proposals for each point. After generating proposals, we extract features for final inference. Our framework is illustrated in Figure \ref{fig:framework}.  

\subsection{Point-based Proposal Generation}

\begin{figure}[bpt]
	\centering
	\includegraphics[width=0.8\linewidth]{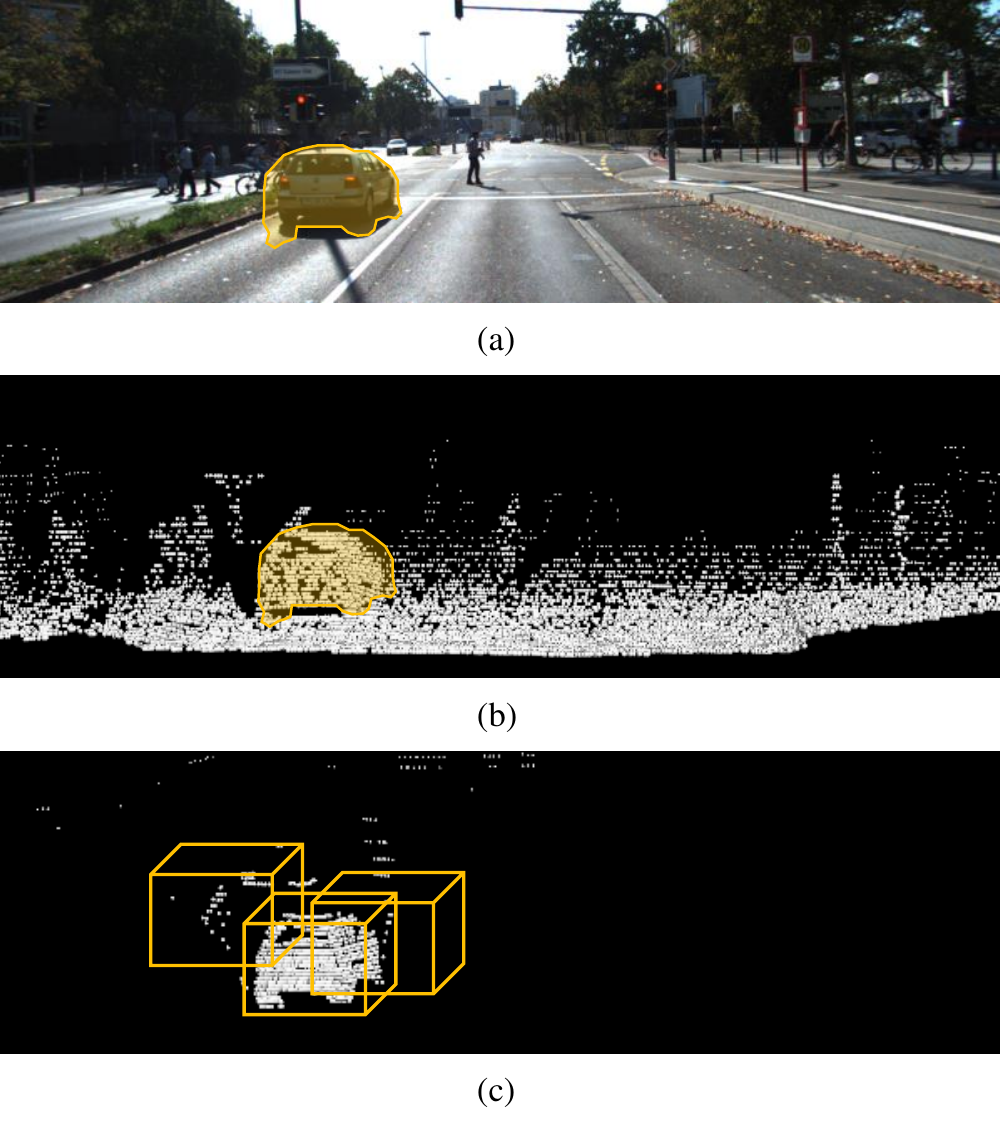}
	\caption{Illustration of point-based proposal generation. (a) Semantic segmentation result on the image. (b) Projected segmentation result on point cloud. (c) Point-based proposals on positive points after NMS.}
	\label{fig:projection}
\end{figure}

The point cloud gathered from LiDAR is unordered, making it nontrivial to utilize the powerful CNN to achieve good performance. Existing methods mainly project point cloud to different views or divide them into voxels, transforming them into a compact structure. We instead choose a more general strategy to seed object proposals based on each point independently, which is the elementary component in the point cloud. Then we process raw point cloud directly. As a result, precise localization information and a high recall are maintained, \ie, achieving a $96.0\%$ recall on KITTI\cite{KITTIDATASET1} dataset. 

\vspace{-0.1in}
\paragraph{Challenges} Albeit elegant, point-based frameworks inevitably face many challenges. For example, the amount of points is prohibitively huge and high redundancy exists between different proposals. They cost much computation during training and inference. Also, ambiguity of regression results and assigning ground-truth labels for nearby proposals need to be resolved.

\begin{figure}[bpt]
	\centering
	\includegraphics[width=0.5\linewidth]{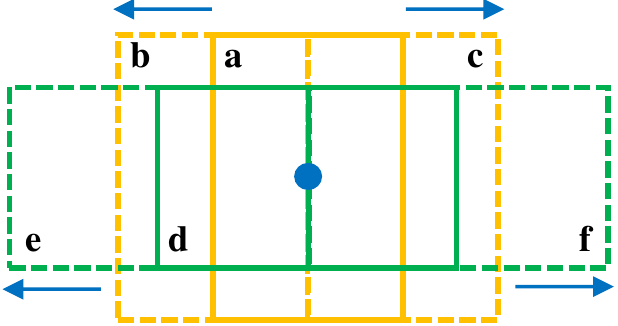}
	\caption{Illustration of proposal generation on each point from BEV. We totally generate 6 proposals based on 2 different anchors with angle of 0 or 90 degree. For each fundamental anchor, we use 3 different shifts along horizontal axis at ratios of -0.5, 0 and 0.5.}
	\label{fig:anchors}
\end{figure}

\vspace{-0.1in}
\paragraph{Selecting Positive Points} The first step is to filter out background points in our framework. We use a 2D semantic segmentation network named subsampling network to predict the foreground pixels and then project them into point cloud as a mask with the given camera matrix to gather positive points. As shown in Figure \ref{fig:projection}, the positive points within a bounding box are clustered. We generate proposals at the center of each point with multiple scales, angles and shift, which is illustrated in Figure \ref{fig:anchors}. These proposals can cover most of the points within a car. 

\vspace{-0.1in}
\paragraph{Reducing Proposal Redundancy} After background point removal, around 60K proposals are left; but many of them are redundant. We conduct non-maximum suppression (NMS) to remove the redundancy. The score for each proposal is the sum of semantic segmentation scores of interior points, making it possible to select proposals with a relatively large number of points. The intersection-over-union (IoU) value is calculated based on the projection of each proposal to the BEV. With these operations, we reduce the number of effective proposals to around 500 while keeping a high recall. 

\begin{figure}[bpt]
	\centering
	\includegraphics[width=0.8\linewidth]{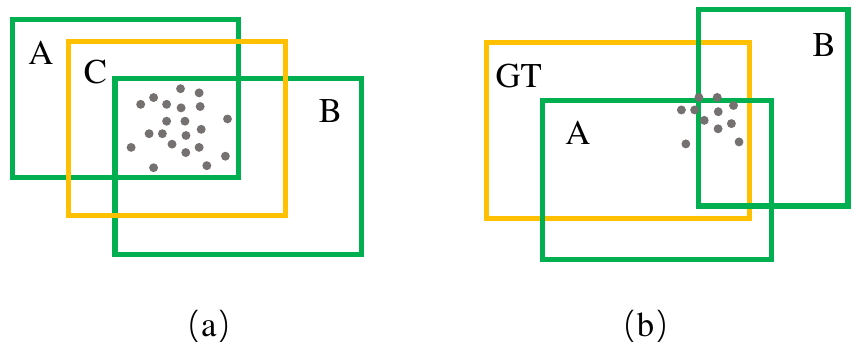}
	\caption{Illustration of paradox situations. (a) Different proposals with the same output. (b) True positive proposal assigned to a negative label. }
	\label{fig:proposal_align}
\end{figure}

\begin{figure*}[bpt]
\centering
\includegraphics[width=1.0\linewidth]{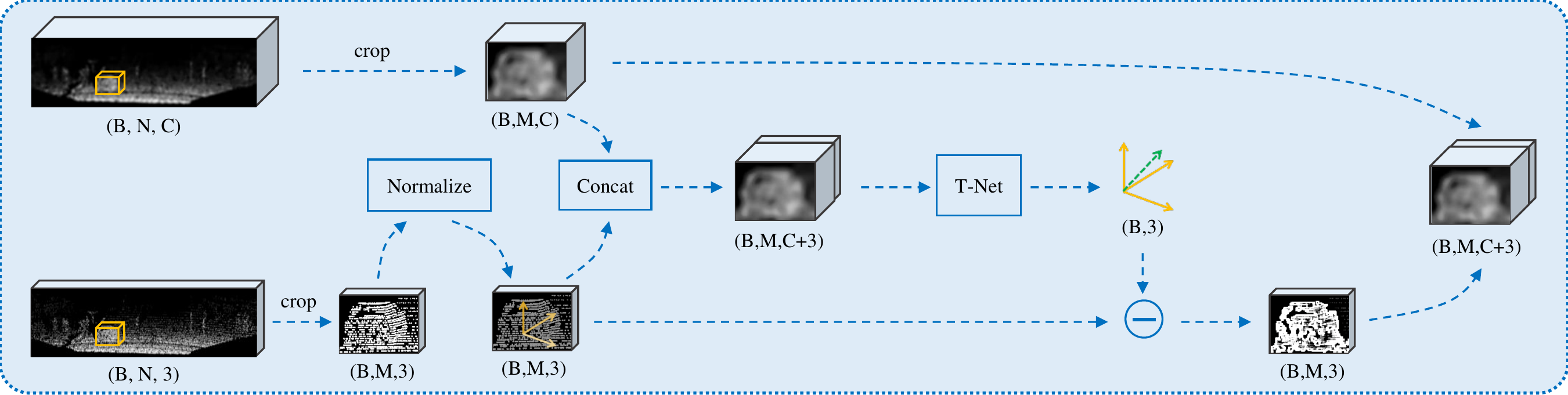}\\
\caption{Illustration of proposal feature generation module. It combines location information and context feature to generate offsets from the centroid of interior points to the center of target instance object. The predicted residuals are added back to the location information in order to make feature more robust to geometric transformation.}
\label{fig:pointspooling}
\end{figure*}

\vspace{-0.1in}
\paragraph{Reduction of Ambiguity}
There are cases that two different proposals contain the same set of points, as shown in Figure \ref{fig:proposal_align}(a). Since the feature for each proposal is produced using the interior points, these proposals thus possess the same feature representation, 
leading to the same classification or regression prediction and yet different bounding box regression results. To eliminate this contradiction, we align these two proposals by replacing their sizes and centers with pre-defined class-specific anchor size and the center of the set of interior points. As illustrated in Figure \ref{fig:proposal_align}(a), the two different proposals A and B are aligned with these steps to proposal C.

Another ambiguity lies on assigning target labels to proposals during training. It is not appropriate to assign positive and negative labels considering only IoU values between proposals and ground-truth boxes, as what is performed in 2D object detector. As shown by Figure \ref{fig:proposal_align}(b), proposal A contains all points within a ground-truth box and overlaps with this box heavily. It is surely a good positive proposal. Contrarily, proposal B is with a low IoU value and yet contains most of the ground-truth points. With the criterion in 2D detector only considering box IoU, proposal B is negative. Note in our point-based settings, interior points are with much more importance. It is unreasonable to consider the bounding box IoU. 

Our solution is to design a new criterion named PointsIoU to assign target labels. PointsIoU is defined as the quotient between the number of points in the intersection area of both boxes and the number of points in the union area of both boxes. According to PointsIoU, both proposals in Figure \ref{fig:proposal_align}(b) are now positive.

\subsection{Network Architecture}

Accurate object detection requires the network to be able to produce correct class label with precise localization information for each instance. As a result, our network needs to be aware of context information to help classification and utilize fine-grained location in raw point cloud. Our network takes entire point cloud as input and produces the feature representation for each proposal. As shown in Figure \ref{fig:framework}, our network consists of a backbone, a proposal feature generation module and a box prediction network.

\begin{figure*}[bpt]
\centering
\begin{tabular}{@{\hspace{0mm}}c}
\includegraphics[width=1.01\linewidth]{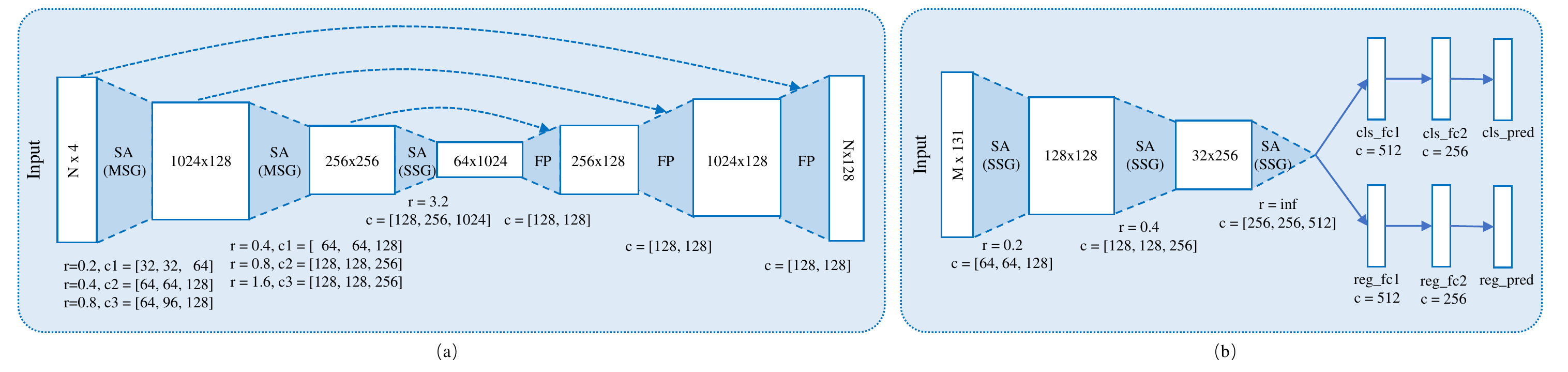}\\
\end{tabular}
\caption{Illustration of two network architectures. (a) Backbone architecture. It takes a raw point cloud $(x, y, z, r)$ as input, and extracts both local and global features for each point by stacking SA layers and FP modules. (b) Bounding-box prediction network. It takes the feature from proposal feature generation module as input and produces classification and regression prediction.}
\label{fig:pred_network}
\end{figure*}

\vspace{-0.1in}
\paragraph{Backbone Network}  The backbone network based on PointNet++ \cite{POINTNET2} takes entire point cloud, with each point parameterized by coordinate and reflectance value, \ie, $([x, y, z, r])$. The network is composed of a number of {\it set\ abstraction} (SA) levels and  {\it feature propagation} (FP) layers, effectively gathering local features from neighboring points and enlarging the receptive field for each point. For $N \times 4$ input points, the network outputs the feature map with size $N \times C$ where each row represents one point. Computation is shared by all different proposals, greatly reducing computation. Since features are generated from the raw points, no voxel projection is needed. The detail of our backbone network is shown in Figure \ref{fig:pred_network}(a).

\vspace{-0.1in}
\paragraph{Proposal Feature Generation} The feature of each proposal has two parts, as shown in Figure \ref{fig:pointspooling}. The first is cropped from the extracted feature map. Specifically, for each proposal, we randomly select $M=512$ points. Then we take corresponding feature vector with size $M \times C$ and denoted it as $F_1$. With the SA and FP operations in PointNet++, these features well capture context information. 

Besides this high-level abstraction, point location is also with great importance. Our second part is the proposal feature $F_2$, the canonized coordinates of the $M$ selected points. Compared with the original coordinates, the canonized ones are more robust to the geometric transformation. We utilize T-Net, which is one type of supervised Spatial Transformation Network (STN) \cite{STN}, to calculate residuals from proposal center to real object center, denoted as $\Delta C_{ctr}$. The input to T-Net is the concatenation of part 1 of proposal feature $F_1$ and the $M$ points' corresponding $XYZ$ coordinates, normalized by subtracting the center coordinates of these $M$ points. As shown in Figure \ref{fig:pointspooling}, part 2 of proposal feature is the canonized coordinates of these $M$ points, calculated by normalized coordinates subtracting the center shift predicted by T-Net. The final proposal feature is the concatenation of $F_1$ and $F_2$.

\vspace{-0.1in}
\paragraph{Bounding-Box Prediction Network}
In this module, for each proposal, we use a small PointNet++ to predict its class, size ratio, center residual as well as orientation. Detailed structure of our prediction network is illustrated in Figure \ref{fig:pred_network}(b). We utilize 3 SA modules with MLP layers for feature extraction. Then average pooling is used to produce the global feature. Two branches for regression and classification is applied. For size ratio, we directly regress the ratio of ground-truth size and the proposal's, parametrized by $(t_l, t_h, t_w)$. We further predict the shift $(t_x, t_y, t_z)$ from refined center by T-Net to ground-truth center. As a result, final center prediction is calculated by $C_{pro} + \Delta C_{ctr} + \Delta C_{ctr}^{\ast}$, where $C_{pro}$, $\Delta C_{ctr}$ and $\Delta C_{ctr}^{\ast}$ denote the center of proposal, prediction of T-Net and shift from bounding-box prediction network, respectively. For heading angle, we use a hybrid of classification and regression formulation following \cite{FPOINTNET}. 
Specifically, we pre-define $N_a$ equally split angle bins and classify the proposal's angle into different bins. Residual is regressed with respect to the bin value. $N_a$ is set to 12 in our experiments.

\vspace{-0.1in}
\paragraph{Loss Functions}
We use a multi-task loss to train our network. The loss function is defined as Eq. \eqref{eq:wholelossfunction}, where $L_{cls}$ is the labeled classification loss. $L_{loc}$ denotes location regression loss, $L_{ang}$ and $L_{cor}$ are angle and corner losses respectively. $s_i$ and $u_i$ are the predicted semantic score and ground-truth label for proposal $i$, respectively. $N_{cls}$ and $N_{pos}$ are the number of proposals and positive samples.
\begin{equation} \label{eq:wholelossfunction}
\begin{aligned}
L_{total} &= \frac{1}{N_{cls}} \sum_i L_{cls} (s_i, u_i) \\
&+ \lambda \frac{1}{N_{pos}} \sum_i  [u_i \geq 1] (L_{loc} + L_{ang} + L_{cor}),
\end{aligned}
\end{equation}
The Iverson bracket indicator function $[u_i \geq 1]$ reaches 1 when $u_i \geq 1$ and 0 otherwise. 
Note that we only determine if each proposal is positive or negative. $L_{cls}$ is simply the softmax cross-entropy loss.

We parameterize a proposal $p$ by its center $(p_x, p_y, p_z)$, size $(p_l, p_h, p_w)$, angle $p_\theta$ and its assigned ground truth box $g$ as $(g_x, g_y, g_z)$, $(g_l, g_h, g_w)$ and $g_\theta$.
Location regression loss is composed of T-Net center estimation loss, center residual prediction loss and size residual prediction loss, expressed as
\begin{equation} \label{eq:regnoangle}
\begin{aligned}
L_{loc} =\ &L_{dis}(t_{ctr}, v_{ctr}) + L_{dis}(t_{ctr}^{\ast}, v_{ctr}^{\ast}) + \\
		  &L_{dis}(t_{size}^{\ast}, v_{size}^{\ast}),
\end{aligned}
\end{equation}
where $L_{dis}$ is the smooth-$l_1$ loss. $t_{ctr}$ and $t_{ctr}^{\ast}$ are predicted center residuals by T-Net and regression network respectively, while $v_{ctr}$ and $v_{ctr}^{\ast}$ are targets for them. $t_{size}^{\ast}$ is the predicted size ratio and $v_{size}^{\ast}$ is the size ratio of ground-truth. The target of our network is defined as
\begin{eqnarray} 
\left \{ \begin{array}{llll}
        v_{ctr}&=& g_k - p_k \textrm{,} & k \in (x,\ y,\ z)\\
       	v_{ctr}^{\ast} &=& g_k - p_k - t_{k}\textrm{,} & k \in (x,\ y,\ z) \\
        v_{size}^{\ast} &=& (g_k - p_k)/p_k \textrm{,} & k \in (l,\ h,\ w)\\
        \end{array} \right.
\end{eqnarray}

Angle loss includes orientation classification loss and residual prediction loss as
\begin{equation} \label{eq:anglereg}
L_{angle} = L_{cls}(t_{a-cls}, v_{a-cls}) + L_{dis}(t_{a-res}, v_{a-res}),
\end{equation}
where $t_{a-cls}$ and $t_{a-res}$ are predicted angle class and residual while $v_{a-cls}$ and $v_{a-res}$ are their targets.   

Corner loss is the distance between the predicted 8 corners and assigned ground truth, expressed as
\begin{equation} \label{eq:cornerloss}
L_{corner} = \sum_{k=1}^8 \norm{P_{vk} - P_{tk}},
\end{equation}
where $P_{vk}$ and $P_{tk}$ are the location of ground-truth and prediction for point $k$. 

\begin{table*}[t]
	\centering 
	\footnotesize
	\begin{tabular}{|c|c||c|c|c||c|c|c||c|c|c|}
	    \hline
	    \multicolumn{1}{|c|}{ \multirow{2}{*}{Method}} & \multicolumn{1}{c||}{ \multirow{2}{*}{Class}} & \multicolumn{3}{|c||}{$AP_{2D} (\%)$} & \multicolumn{3}{|c||}{$AP_{BEV} (\%)$}& \multicolumn{3}{|c|}{$AP_{3D} (\%)$} \\ \cline{3-11}
	    \multicolumn{1}{|c|}{} & \multicolumn{1}{c||}{} & \multicolumn{1}{|c|}{Easy} & \multicolumn{1}{|c|}{Moderate} & \multicolumn{1}{|c||}{Hard} & \multicolumn{1}{|c|}{Easy} & \multicolumn{1}{|c|}{Moderate} & \multicolumn{1}{|c||}{Hard} & \multicolumn{1}{|c|}{Easy} & \multicolumn{1}{|c|}{Moderate} & \multicolumn{1}{|c|}{Hard} \\
	    \hline
	    \hline
		MV3D \cite{MV3D} & \multirow {6}{*}{Car} & N/A & N/A & N/A & 86.02 & 76.90 & 68.49 & 71.09 & 62.35 & 55.12\\
		AVOD \cite{AVOD} & {} & 89.73 & 88.08 & 80.14 & 86.80 & \bf 85.44 & 77.73 & 73.59 & 65.78 & 58.38 \\
	    VoxelNet \cite{VOXELNET} & {} & N/A & N/A & N/A & \bf 89.35 & 79.26 & 77.39 & 77.47 & 65.11 & 57.73 \\
		F-PointNet \cite{FPOINTNET} & {} & \bf 90.78 & \bf 90.00 & 80.80 & 88.70 & 84.00 & 75.33 & 81.20 & 70.39 & 62.19 \\ 
		AVOD-FPN \cite{AVOD} & {} & 89.99 & 87.44 & 80.05 & 88.53 & 83.79 & \bf 77.90 & \bf 81.94 & 71.88 & \bf 66.38 \\
		Ours & {} & 90.20 & 89.30  & \bf 87.37 & 86.93 & 83.98 & 77.85 & 79.75 & \bf 72.57 & 66.33 \\
		\hline
		\hline
		AVOD \cite{AVOD} & \multirow {5}{*}{Pedestrian} & 51.64 & 43.49 & 37.79 & 42.51 & 35.24 & 33.97 & 38.28 & 31.51 & 26.98 \\
	    VoxelNet \cite{VOXELNET} & {} & N/A & N/A & N/A & 46.13 & 40.74 & 38.11 & 39.48 & 33.69 & 31.51 \\
		F-PointNet \cite{FPOINTNET} & {} & \bf 87.81 & \bf 77.25 & \bf 74.46 & 58.09 & 50.22 & 47.20 & 51.21 & \bf 44.89 & 40.23 \\ 
		AVOD-FPN \cite{AVOD} & {} & 67.32 & 58.42 & 57.44 & 58.75 & 51.05 & \bf 47.54 & 50.80 & 42.81 & 40.88 \\
		Ours & {} & 73.28 & 63.07 & 56.71 & \bf 60.83 & \bf 51.24 & 45.40 & \bf 56.92 & 44.68 & \bf 42.39 \\
		\hline
		\hline
		AVOD \cite{AVOD} & \multirow {5}{*}{Cyclist} & 65.72 & 56.01 & 48.89 & 63.66 & 47.74 & 46.55 & 60.11 & 44.90 & 38.80 \\
	    VoxelNet \cite{VOXELNET}& {} & N/A & N/A & N/A & 66.70 & 54.76 & 50.55 & 61.22 & 48.36 & 44.37 \\
		F-PointNet \cite{FPOINTNET} & {} & \bf 84.90 & \bf 72.25 & \bf 65.14 & 75.38 & \bf 61.96 & \bf 54.68 & \bf 71.96 & \bf 56.77 & \bf 50.39 \\ 
		AVOD-FPN \cite{AVOD} & {} & 68.65 & 59.32 & 55.82 & 68.09 & 57.48 & 50.77 & 64.00 & 52.18 & 46.61 \\
		Ours & {} & 82.90 & 65.28 & 57.63 & \bf 77.10 & 58.92 & 51.01 & 71.40 & 53.46 & 48.34 \\
		\hline
	\end{tabular}\vspace{0.1cm}
	\caption{Performance on KITTI test set for both Car, Pedestrian and Cyclists.}\label{tab:mainkitti}
\end{table*}

\section{Experiments}
We evaluate our method on the widely employed KITTI Object Detection Benchmark \cite{KITTI3DBENCHMARK}. There are 7,481 training images / point clouds and 7,518 test images / point clouds in three categories of Car, Pedestrian and Cyclist. We use the most widely used average precision (AP) metric to compare different methods. During evaluation, we follow the official KITTI evaluation protocol -- that is, the IoU threshold is 0.7 for class Car and 0.5 for Pedestrian and Cyclist. 

\subsection{Implementation Details}
Following \cite{VOXELNET,AVOD}, we train two networks, one for car and the other for both pedestrian and cyclist.

\vspace{-0.1in}
\paragraph{Positive Points Selection}
For semantic segmentation, which is used for foreground point selection, we use the Deeplab-v3 \cite{DEEPLAB} based on X-ception \cite{XCEPTION} network trained on Cityscapes \cite{CITYSCAPE} due to the lack of semantic segmentation annotation in KITTI dataset. Pixels labeled as {\tt car}, {\tt rider}, {\tt person} or {\tt bicycle} are regarded as foreground. There is no class {\tt cyclist} in Cityscapes. We thus combine bicycle and rider segmentation masks for {\tt cyclist}. After getting positive masks, we project these pixels back to the 3D points and acquire foreground points. To align network input, we randomly choose $N$ points per point cloud from foreground. If positive point number is less than $N$, we pad to $N$ points from the negative points. During training and testing, we set $N = 10K$ for the car detection model and $N = 5K$ for the pedestrian and cyclist model. 

\vspace{-0.1in}
\paragraph{Anchor Sizes}
We define the anchor size $(l^a=3.9, h^a=1.6, w^a=1.6)$ for car detection model, and sizes $(l^a=1.6, h^a=1.6, w^a=0.8)$ and $(l^a=0.8, h^a=1.6, w^a=0.8)$ for the other model. For each size, we use 2 different angles, and 3 shifts, yielding $k=6$ proposals centered at the location of each point, as illustrated in Figure \ref{fig:anchors}. 

\vspace{-0.1in}
\paragraph{Training Parameters}
During training, we use ADAM \cite{AdamOptimizer} optimizer with an initial learning rate of 0.001 for the first 90 epochs and then decay the learning rate by 0.1 in every 10 epochs. We train 120 epochs in total. Each batch consists of 8 point clouds evenly distributed on 4 GPU cards. For each input point cloud, we sample 64 proposals, with a ratio of 1:3 for positives and negatives. Our implementation is based on Tensorflow \cite{Tensorflow}. During training the car model, a proposal is considered positive if its PointsIoU with a certain ground-truth box is higher than 0.55 and negative if its PointsIoU is less than 0.55 with all ground-truth boxes. The positive and negative PointsIoU thresholds are 0.5 and 0.5 for the pedestrian and cyclist model.

\vspace{-0.1in}
\paragraph{Data Augmentation}
Data augmentation is important to prevent overfitting. We have used four point cloud augmentation methods following \cite{VOXELNET}. First, all instance objects with their interior points are randomly transformed respectively. For each bounding box, we randomly rotate it by a uniform distribution $\Delta \theta_1 \in [- \pi / 3, + \pi / 3]$ and randomly add a translation ($\Delta x, \Delta y, \Delta z$) to its $XYZ$ value as well as its interior points. Second, each point cloud is flipped along the $x$-axis in camera coordinate with probability 0.5. Third, we randomly rotate each point cloud around $y$-axis (up orientation) by a uniformly distributed random variable $\Delta \theta_2 \in [- \pi / 4, + \pi / 4]$. Finally, we apply a global scaling to point cloud with a random variable drawn from uniform distribution $[0.9, 1.1]$. During training, we randomly use one of these four augmentations for each point cloud.

\vspace{-0.1in}
\paragraph{Post-process}
After predicting the bounding box for each proposal, we use a NMS layer with IoU threshold 0.01 to remove overlapped detection. The input scores to NMS are classification probabilities, and the predicted bounding boxes are projected to BEV before performing NMS.

\begin{figure}[t]
\centering
\includegraphics[width=1.0\linewidth]{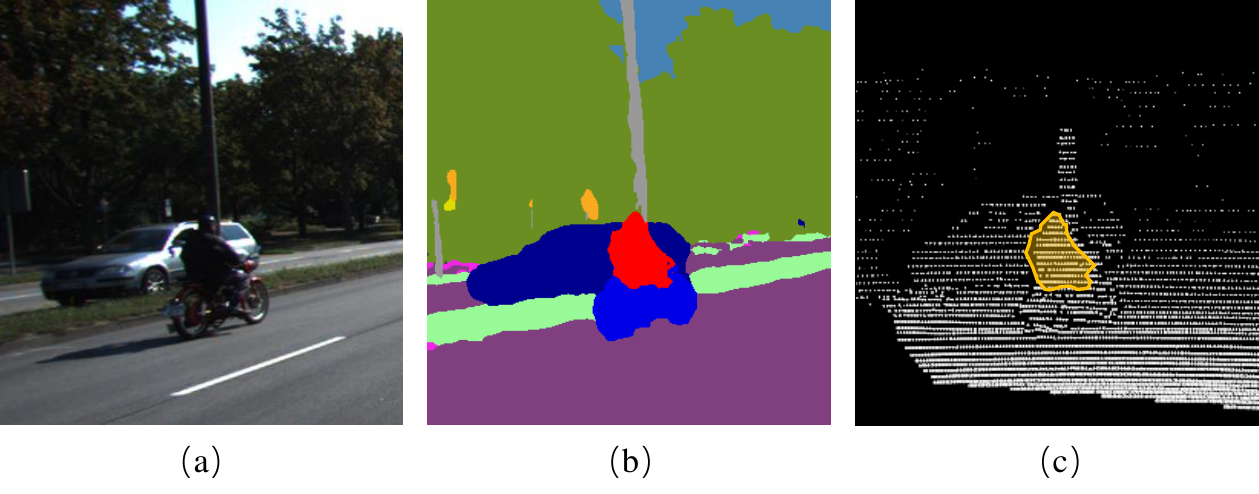}
\caption{Illustration of failure cases in Cyclist detection. In this case, we can only get points labeled as rider, which leads to the detection result tending to be Pedestrian rather than Cyclist.}
\label{fig:cyclist}
\end{figure}

\subsection{Evaluation on KITTI Test Set}
For evaluation on the test set, we train our model on our own split train/val set at a ratio of 4:1. The performance of our method and its comparison with previous state-of-the-arts is listed in Table \ref{tab:mainkitti}. Compared to F-PointNet, our model greatly improves the detection accuracy on the hard set by $6.75\%$, $2.52\%$, and $4.14\%$ on 2D, BEV and 3D respectively, which indicates that our model can overcome the limitation of 2D detector on cluttered objects. 

\begin{figure*}[t]
	\centering
	\begin{tabular}{@{\hspace{0mm}}c}
		\includegraphics[width=1.0\linewidth]{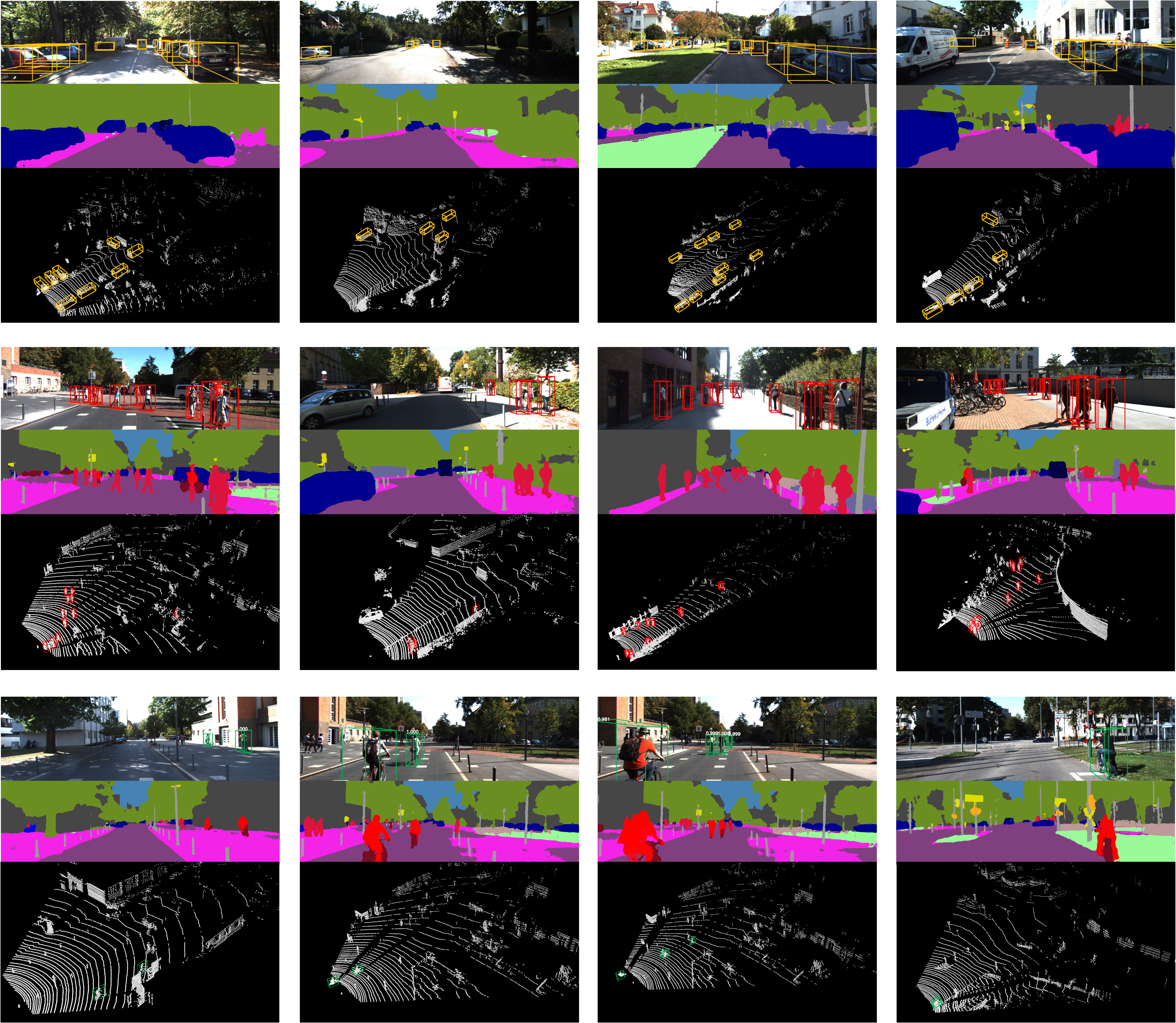}\\
	\end{tabular}
	\caption{Visualizations of our results on KITTI test set. In each image, a box detected as car, pedestrian or cyclist is in yellow, red or green respectively. The top row in each image is 3D object detection results projected on to the RGB images. The middle is 2D semantic segmentation results of images. The bottom is 3D object detection results in the LIDAR phase.}
	\label{fig:results}
\end{figure*}

Compared to multi-view based methods \cite{AVOD,MV3D}, ours performs better in Pedestrian prediction by a large margin of $6.12\%$, $1.87\%$, and $1.51\%$ on the easy, moderate and hard levels separately. So our strategy is good at detecting dense objects, such as a crowd of people. Compared to VoxelNet\cite{VOXELNET}, our model shows superior performance in all classes. We present several detection results in Figure \ref{fig:results}, where many difficult cases are decently dealt with. 

We note that one current problem, due to the dataset limitation, is the absence of class {\tt cyclist} in Cityscapes dataset. So it is hard to select foreground points of cyclists, leading to relatively weak performance on this class. Several imperfect semantic masks are shown in Figure \ref{fig:cyclist}.

\subsection{Ablation Studies}
For ablation studies, we follow VoxelNet \cite{VOXELNET} to split the official training set into a {\it train} set of 3,717 images and a {\it val} set of 3,769 images. Images in train/val set belonging to different video clips. All ablation studies are conducted on the {\tt car} class due to the relatively larger amount of data to make system run stably. 

%  Whether mask table
\begin{table}[t]
	\centering \addtolength{\tabcolsep}{-1pt}
	\footnotesize
	\begin{tabular}{|c|ccc|}
	    \hline
	    Mask & Easy & Moderate & Hard \\
	    \hline
	    No Mask (Random Choice) & 71.1 & 55.5 & 49.7 \\
	    No Mask (Sliding Window) & 76.4 & 60.0 & 51.9 \\
	    3D (Random Choice) & 72.8 & 63.6 & 60.9 \\
	    3D (Sliding Window) & 77.4 & 69.7 & 65.9 \\
	    2D & \bf 84.1 & \bf 76.4 & \bf 75.3 \\
		\hline
	\end{tabular}\vspace{0.3cm}
	\caption{3D object detection AP on KITTI val set. ``Random Choice" means randomly sampling points inside a point cloud. ``Sliding Window" stands for taking one part of point cloud as input each time and ensembling these detection results. ``No Mask" means directly using all points in a point cloud as positive points. ``3D" means directly using 3D segmentation module as the subsampling method. ``2D" means using 2D image segmentation as the subsampling method.}
	\label{tab:whetherseg}
\end{table}

% Why points sampling
\begin{table}[t]
	\centering \addtolength{\tabcolsep}{-1pt}
	\footnotesize
	\begin{tabular}{|c|c|ccc|}
		\hline
		Points IoU & Context Feature & Easy & Moderate & Hard \\
		\hline
		- & - & 61.4 & 50.8 & 44.9 \\
		$\surd$ & - & 68.1 & 56.3 & 52.2 \\
		$\surd$ & $\surd$ & \bf 84.1 & \bf 76.4 & \bf 75.3 \\
		\hline
	\end{tabular}\vspace{0.3cm}
	\caption{3D object detection AP on KITTI val set. A tick in the ``Points IoU" item means we use Points IoU. Otherwise, we use original bounding box IoU for assigning positive and negative proposals as an alternative. A tick in the ``Context Feature" item means we use the PointNet++ backbone to extract context feature for the whole point cloud.}
	\label{tab:whetherpointsiou}
\end{table}

\vspace{-0.1in}
\paragraph{Effect of Subsampling Network}
Subsampling network in our model is to eliminate complicated background points so as to reduce computation complexity. We conduct experiments on using 2D segmentation and 3D segmentation as point selection masks. Here, we choose 2D segmentation rather than 2D detection since segmentation results are with much higher quality and more useful on cluttered scenes. As listed in Table \ref{tab:whetherseg}, 2D semantic segmentation is the best choice empirically for our model. 

However, if we use 3D semantic segmentation module instead, either sampling points or taking one part of point cloud as input each time by sliding window is needed, due to the GPU memory limitation. Our observation is that results with 3D segmentation are even with lower quality than those from 2D segmentation. We also compare with the baseline solution without masking foreground points, denoted as 'No Mask'. Still, this baseline performs notably worse than our full design, manifesting the importance of subsampling network.

% Result on valiadation set
\begin{table}[t]
	\centering \addtolength{\tabcolsep}{-1pt}
	\footnotesize
	\begin{tabular}{|c|ccc|}
	    \hline
	    Benchmark & Easy & Moderate & Hard \\
	    \hline
	    Car (3D Detection) & 84.1 & 76.4 & 75.3 \\
	    Car (BEV) & 88.3 & 86.4 & 84.6 \\
	    \hline
	    Pedestrian (3D Detection) & 69.6 & 62.3 & 54.6 \\
	    Pedestrian (BEV) & 72.4 & 67.8 & 59.7 \\
		\hline
		Cyclist (3D Detection) & 81.9 & 57.1 & 54.6 \\
		Cyclist (BEV) & 84.3 & 61.8 & 57.7 \\
		\hline
	\end{tabular}\vspace{0.3cm}
	\caption{3D object detection AP and BEV detection AP on KITTI val set of our model for cars, pedestrians and cyclists.}
	\label{tab:kittival}
\end{table}

\vspace{-0.1in}
\paragraph{Effect of PointsIoU Sampling and Context Feature}
PointsIoU sampling is used in the training phase as the criterion to assign positive and negative proposals. As shown in Table \ref{tab:whetherpointsiou}, it significantly promotes our model's performance by $7.3\%$ in hard cases, compared with trivially using bounding box IoU. It is helpful also in suppressing false positive proposals as well as false negative proposals. PointNet++ backbone can extract the context feature for the whole point cloud, which is beneficial to 3D detection. AP in Table \ref{tab:whetherpointsiou} demonstrates its ability, which increases by around $15\%$ on easy set, $20\%$ on both moderate and hard sets.

\vspace{-0.1in}
\paragraph{Effects of Proposal Feature}
As shown in Figure \ref{fig:pointspooling}, our proposal feature is composed of high-level abstraction feature and canonized coordinates. We investigate the effect of using different components in the point pooling feature. As shown in Table \ref{tab:whethertnet}, combination of high-level feature and canonized coordinates enhance the model's capability greatly and is thus adopted in our final structure.

% Why T-Net
\begin{table}[t]
	\centering \addtolength{\tabcolsep}{-1pt}
	\footnotesize
	\begin{tabular}{|c|c|c|c|}
	    \hline
	    high-level & normalized & canonized & accuracy \\
	    \hline
	    - & $\surd$ & - & 56.3 \\
	    \hline
	    $\surd$ & - & - & 74.2 \\
	    \hline
	    $\surd$ & $\surd$ & - & 75.0 \\
	    \hline
	    $\surd$ & - & $\surd$ & \bf 76.4 \\
		\hline
	\end{tabular}\vspace{0.3cm}
	\caption{Effect of using different components of proposal feature. A tick in ``high-level" item means using points feature with high-level abstraction and context information from the backbone network. A tick in ``normalized" item means we simply use the coordinates normalized by points center as the second part of proposal feature. Confirmation on the ``canonized" item means the second part proposal feature is the canonized $XYZ$ value instead of the normalized $XYZ$.} 
	\label{tab:whethertnet}
\end{table}

\vspace{-0.1in}
\paragraph{Result on Validation} 
Our results on KITTI validation set including ``Car", ``Pedestrian" and ``Cyclist" are shown in Table \ref{tab:kittival}, with all components included. Our results on validation set compared to other state-of-the-art methods are listed in Table \ref{tab:kittivalcompare}. Compared to F-PointNet, our model performs notably better on small objects with only a few points, which brings great benefit to detection on hard set. Moreover, compared to multi-view-based methods like \cite{AVOD}, our method also yields intriguing 3D object detection performance. It is because our model is more sensitive to structure and appearance details.

% Results on validation set compared to other method
\begin{table}[t]
	\centering \addtolength{\tabcolsep}{-1pt}
	\footnotesize
	\begin{tabular}{|c|ccc|}
		\hline
		Method & Easy & Moderate & Hard \\
		\hline
		VoxelNet \cite{VOXELNET} (3D) & 81.97 & 65.46 & 62.85 \\
		VoxelNet \cite{VOXELNET} (BEV) & \bf89.60 & 84.81 & 78.57 \\
		\hline
		F-PointNet \cite{FPOINTNET} (3D) & 83.76 & 70.92 & 63.65 \\
		F-PointNet \cite{FPOINTNET} (BEV) & 88.16 & 84.02 & 76.44 \\
		\hline
		AVOD \cite{AVOD} (3D) & \bf 84.41 & 74.44 & 68.65 \\
		AVOD \cite{AVOD} (BEV) & - & - & - \\
		\hline
		Ours (3D) & 84.1 & \bf 76.4 & \bf 75.3 \\
		Ours (BEV) & 88.3 & \bf 86.4 & \bf 84.6 \\
		\hline
	\end{tabular}\vspace{0.3cm}
	\caption{3D and BEV detection AP on KITTI val set of our model for ``Car" compared to other state-of-the-art methods.}
	\label{tab:kittivalcompare}
\end{table}

\section{Concluding Remarks}
We have proposed a new method operating on raw points. We seed each point with proposals, without loss of precious localization information from point cloud data. Then prediction for each proposal is made on the proposal feature with context information captured by large receptive field and point coordinates that keep accurate shape information. Our experiments have shown that our model outperforms state-of-the-art 3D detection methods in hard set by a large margin, especially for those high occlusion or crowded scenes.

{\small
\bibliographystyle{ieee}
\bibliography{egbib}
}

\end{document}